# EV-PINN: A Physics-Informed Neural Network for Predicting Electric Vehicle Dynamics

Hansol Lim, Jee Won Lee, Jonathan Boyack, and Jongseong Brad Choi, *Member, IEEE*

*Abstract*—An onboard prediction of dynamic parameters (e.g. Aerodynamic drag, rolling resistance) enables accurate path planning for EVs. This paper presents EV-PINN, a Physics-Informed Neural Network approach in predicting instantaneous battery power and cumulative energy consumption during cruising while generalizing to the nonlinear dynamics of an EV. Our method learns real-world parameters such as motor efficiency, regenerative braking efficiency, vehicle mass, coefficient of aerodynamic drag, and coefficient of rolling resistance using automatic differentiation based on dynamics and ensures consistency with ground truth vehicle data. EV-PINN was validated using 15 and 35 minutes of in-situ battery log data from the Tesla Model 3 Long Range and Tesla Model S, respectively. With only vehicle speed and time as inputs, our model achieves high accuracy and generalization to dynamics, with validation losses of 0.002195 and 0.002292, respectively. This demonstrates EV-PINN's effectiveness in estimating parameters and predicting battery usage under actual driving conditions without the need for additional sensors.

## I. INTRODUCTION

Electric vehicles (EVs) are increasingly adopted due to environmental concerns, technological advancements, and supportive government policies, driving significant market growth with projections of EV sales reaching 145 million by 2030 [1]. As EV adoption accelerates, effective energy management becomes crucial for optimizing energy consumption, which is essential for maximizing the efficiency and performance. Various optimization techniques have been explored to address these challenges.

A comprehensive survey on EV battery optimization by Acar et al. [2] discusses traditional methods like dynamic programming, model predictive control, and heuristic algorithms, highlighting their significant limitations, particularly in real-time EV operations. Traditional optimization methods often rely on detailed models of vehicle dynamics and battery behavior, which can be complex and computationally expensive to develop. Dynamic programming optimizes sequential decisions but is faced by what we call "curse of dimensionality," where computational demands increase exponentially with more variables. These methods are expensive and hence unsuitable for real-time use in high-dimensional systems like EVs and

Hansol Lim, Jee Won Lee, Jonathan Boyack, and Jongseong Brad Choi are with Mechanical Engineering, State University of New York, Stony Brook, Stony Brook, NY 11794, USA (e-mail: hansol.lim@stonybrook.edu; jeewon.lee@stonybrook.edu; jonathan.boyack@stonybrook.edu; jongseong.choi@stonybrook.edu).

This work supported by the National Research Foundation of Korea 14 CHOI et al. (NRF) grant funded by the Korea government (MSIT) (No. 2022M1A3C2085237). (Corresponding author: Jongseong Brad Choi).

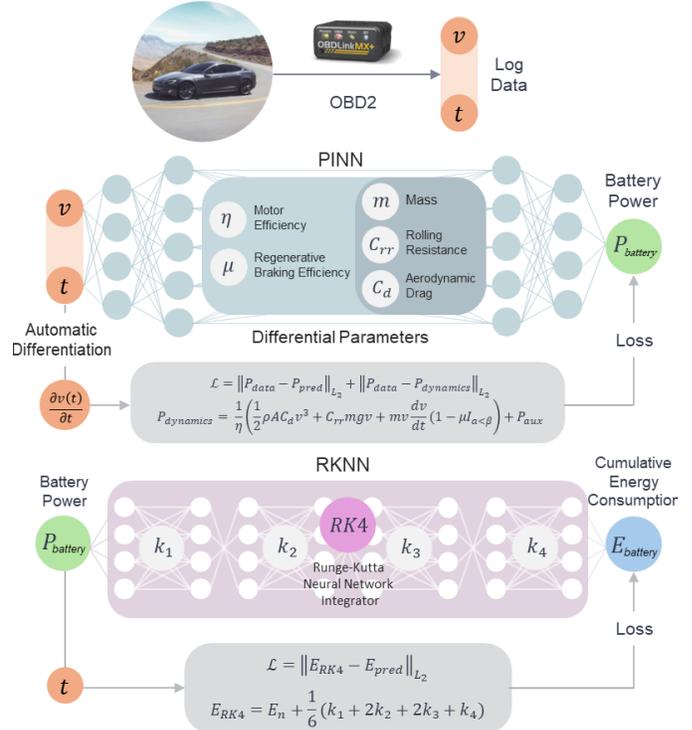

Fig. 1. EV-PINN overview

it may lead to suboptimal performances.

Current MPC optimization models often use linear approximations for nonlinear systems, leading to inaccurate predictions especially under external disturbances that differ significantly from assumptions. Heuristic methods like genetic algorithms and particle swarm optimization handle nonlinearity and complex constraints but they require significant computational resources which may be insufficient for the diverse, dynamic scenarios in real-world EV operations. These limitations underscore the need for more advanced and adaptable optimization techniques that can better handle the complexities and dynamics of EV energy management.

Recent approaches have attempted to address these challenges. Zhang [3] used optimal control techniques like convex optimization and second-order conic relaxation to efficiently manage EV charging and discharging loads, enhancing grid stability and reducing operational costs. Zhu et al. [4] applied Long Short-Term Memory (LSTM) networks to optimize EV charging decisions by taking its advantages in handling time-series data. This approach has demonstrated significant improvements in reducing charging costs and enhancing the load management of the power grid [4]. While the methods were effective, they depend on extensive sensor data and expensive computation, and may not always accurately capture the dynamic nature of EV operation.

Physics-Informed Neural Networks (PINNs), introduced by Raissi et al. [5], offer a promising alternative by integrating physical laws directly into neural network training. PINN's potential for predicting PDE solutions gives opportunities for a more versatile and reliable framework for prediction of nonlinear models. It also seems particularly useful in scenarios where the available data is sparse or incomplete. More explanation on PINNs will be provided in Reated Work section of the paper.

That said, we present a novel application of PINN for modeling EV dynamics using log data from the Tesla Model 3 Long Range and Tesla Model S. Our model estimates real-world conditions such as motor efficiency, regenerative braking efficiency, vehicle mass, aerodynamic drag, and rolling resistance. By incorporating these parameters, the proposed method aims to provide more precise predictions of battery consumption with minimal sensor inputs for EVs, just using vehicle speed and time.

## II. RELATED WORKS

### A. Neural Networks as Universal Approximators

Neural networks excel in approximating a wide range of functions, supported by the Universal Approximation Theorem, which states that a feedforward neural network with a single hidden layer can approximate any continuous function with sufficient neurons. [6]. This capability arises from their architecture, where each neuron takes a weighted sum of inputs, applies a non-linear activation function, and passes the result to the next layer. By combining linear and nonlinear transformations, neural networks transform input data into a latent space, which, in a sense, encodes and decodes inputs to outputs.

Consider a feedforward neural network with one hidden layer. The output of the network $f(x)$, given an input $x \in \mathbb{R}^n$, can be expressed as:

$$(x) = \sum_{i=1}^{m} \beta_i \sigma \left( \sum_{j=1}^{n} \alpha_{ij} x_j + b_i \right) + c \quad (1)$$

Where $\sigma(\cdot)$ is the activation function, $m$ is the number of neurons in the hidden layer, $\alpha_{ij}$ and $\beta_i$ are weights for the input and hidden layers, respectively, $b_i$ is the bias term for hidden layer, and $c$ is the bias term for output layer.

The Universal Approximation Theorem asserts that for any continuous function $g(x)$ defined on a compact subset $K$ of $\mathbb{R}^n$ and for any $\epsilon > 0$, there exists a neural network with a finite number of neurons $m$ such that:

$$\sup_{x \in K} |f(x) - g(x)| < \epsilon \quad (2)$$

This means that the network can approximate $g(x)$ as closely as desired, given enough neurons and appropriate weights.

### B. Physics-Informed Neural Networks

PINNs extend traditional neural networks by integrating known physical laws into the learning process. This process enables solutions that are both data-driven and consistent with fundamental physics principles. Many natural and engineering systems are governed by Partial Differential Equations (PDEs), such as the Navier-Stokes equations for fluid flow and the Fourier heat equation for heat conduction. Traditionally, solving these PDEs involves numerical methods like finite element method or finite difference method which can be computationally intensive and prone to discretization errors.

While neural networks are adept at approximating complex functions, purely data-driven models can produce solutions that violate physical laws, especially when trained on limited or noisy data. Moreover, models trained on huge datasets cam suffer from overfitting issues where the network fails to generalize and cannot adapt to new situations. To address these limitations, PINNs offer opportunities for generalizable networks that respect fundamental physics by incorporating governing PDEs directly into the neural network's training process [5].

Consider a physical system described by a PDE:

$$\mathcal{N}(u(x,t)) = f(x,t) \quad (3)$$

Where $\mathcal{N}$ is a nonlinear differential operator, $u(x,t)$ is the unknown solution we want to learn, and $f(x,t)$ is a known function. The nonlinear differential operator (neural network) is used to approximate $u(x,t)$, and the loss function $\mathcal{L}_{PINN}$ is formulated as:

$$\mathcal{L}_{PINN} = \sum_{i=1}^{N} \|u(x_i,t_i) - u_{data}(x_i,t_i)\|^2 + \lambda \sum_{j=1}^{M} \|\mathcal{N}(u(x_j,t_j)) - f(x_i,t_i)\|^2 \quad (4)$$

Where $\lambda$ is the weighting parameter that balances the influences of the governing physics to the data.

### C. PINN-Driven Modeling of Complex Systems

A good example of a PINN application is in the field of robotics. Dynamic systems are governed by a system of nonlinear ODEs. For example, the dynamics of a multi-link robot arms are typically described by the following system [7]:

$$M(q)\ddot{q} + V(q,\dot{q}) + G(q) = \tau \quad (5)$$

Where $q$ is generalized coordinates vector (angles), $M(q)$ is the inertia matrix, $V(q,\dot{q})$ represents Coriolis and centrifugal forces, $G(q)$ represents gravitational forces, and $\tau$ is the generalized applied force (torque).

These ODE describes how applied forces affect the robot's joint motion. However, the problem is that real-time solutions are computationally intensive. Traditional numerical methods can be slow or require simplifications that reduce model accuracy.

Nicodemus et al. [8] showed that using PINNs as surrogate models for controlling 6-DOF robot arms can significantly reduce computation time while preserving the accuracy required for real-time control.

$$\mathcal{L}_{manipulator} = \mathcal{L}_{data} + \lambda_1 \mathcal{L}_{dynamics} + \lambda_2 \mathcal{L}_{constraints} \quad (6)$$

This loss formulation allows the PINN to learn a model respecting both Lagrangian dynamics of robot arm. The model results were more robust and more effective against disturbances than traditional methods.

In case of autonomous driving, accurate and reliable vehicle dynamics models are essential for ensuring safe navigation under a wide range of road conditions. Another study by Kim et al. [9], enhanced control strategies for autonomous vehicles (motorcycle) by integrating the Pacejka tire model and a dynamic bicycle model into a PINN framework to predict vehicle behavior under various friction conditions, like icy or wet roads.

$$m(\dot{v}_x + v_y\dot{\psi}) = F_x \quad (7)$$

$$m(\dot{v}_y + v_x\dot{\psi}) = F_y \quad (8)$$

$$I_z\ddot{\psi} = M_z \quad (9)$$

Where $v_x$ and $v_y$ are longitudinal and lateral velocities, $\psi$ is the yaw angle of the vehicle, $F_x$ and $F_y$ are the longitudinal and lateral forces acting on the vehicle, $M_z$ is the yaw moment, $m$ is the mass of the vehicle, and $I_z$ is the moment of inertia about the vehicle's vertical axis.

The PINN approximates the vehicle's state over time $s(t) = [v_x, v_y, \psi, \dot{\psi}]^T$ with the following loss function:

$$\mathcal{L}_{vehicle} = \mathcal{L}_{data} + \lambda \sum_{j=1}^{M} \left\| \mathcal{F}_{residuals}\left(\hat{s}(t_j), u(t_j)\right) \right\|^2 \quad (10)$$

By implementing PINNs into the autonomous driving framework, they were able to accurately capture lateral tire forces, adapt to changing road friction, and mitigate risks like skidding or loss of control.

As shown in the studies above, PINNs effectively combine neural networks and dynamics to approximate real-world states. By integrating empirical data with physical laws, PINNs allow prediction of accurate and robust models that capture complex behaviors while respecting the underlying physics.

## III. METHODOLOGY

### A. Overview

To train EV-PINN, we collected data from Tesla Model 3 Long Range and Tesla Model S, derived battery power dynamics and incorporated them as network losses. After optimizing the hyperparameters, we predicted instantaneous battery power and then used a Runge-Kutta Neural Network to predict cumulative energy consumption.

### B. Log Data Collection

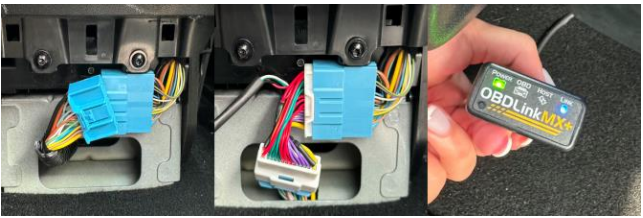

Fig. 2. OBD2 scanner installation on test vehicles

Log data for the Tesla Model 3 Long Range and Model S were collected via OBD2 scanner. The collected data included vehicle speed ($v$), time ($t$), battery voltage ($V$), and battery current ($I$). To minimize external variables, log data were collected in a controlled environment: flat roads, no wind, sunny day, and no traffic.

Ground Truth data for battery power was calculated with Ohm's Law using log data voltage and current.

$$P_{battery,i} = I_i V_i \quad (11)$$

### C. Electric Vehicle Dynamics

In this section, we derive the key ODE that governs the dynamics of the electric vehicle (EV) battery power consumption.

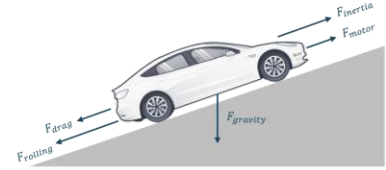

Fig. 3. Free Body Diagram of EV during motion

Following forces describe the dynamics of the EV:

$$F_{drag} = \frac{1}{2}\rho A C_d v^2 \quad (12)$$

$$F_{rolling} = C_{rr} mg \cos(\theta) \quad (13)$$

$$F_{gravity} = mg \sin(\theta) \quad (14)$$

$$F_{inertia} = m\frac{dv}{dt} \quad (15)$$

$$F_{motor} = \frac{\eta(P_{battery} - P_{aux})}{v} \quad (16)$$

Equations (12), (13), (14), (15), and (16) represent aerodynamic drag force, rolling resistance force, gravitational force, inertia force, and battery motor force, respectively. $\rho$ is the air density, $A$ is the frontal area of the vehicle, $C_d$ is the coefficient of aerodynamic drag, $v$ is the velocity of the vehicle, $C_{rr}$ is the coefficient of rolling resistance, $m$ is the mass of the vehicle, $g$ is the acceleration due to gravity, $\theta$ is the road incline angle and $\frac{dv}{dt}$ is the acceleration of the vehicle. $\eta$ is the motor efficiency, $P_{aux}$ is the auxiliary power consumption for Tesla Model 3 LR and Model S and is found to be $\approx 1100W$ and $\approx 390W$, respectively, and was derived from the idle state of the vehicle ($v = 0$, $\frac{dv}{dt} = 0$).

The total force required to move the vehicle is given by the sum of these forces:

$$F_{motor} = F_{drag} + F_{rolling} + F_{gravity} + F_{inertia} \quad (17)$$

By equating the motor force to the sum of the opposing forces, we get:

$$\frac{\eta(P_{battery} - P_{aux})}{v} = \frac{1}{2}\rho A C_d v^2 + C_{rr} mg\cos(\theta) + mg\sin(\theta) + m\frac{dv}{dt} \quad (18)$$

Letting $\theta = 0$ (flat roads) and rearranging to solve for $P_{battery}$, we obtain the governing nonlinear ODE for EV battery power consumption:

$$P_{battery} = \frac{1}{\eta}\left(\frac{1}{2}\rho A C_d v^3 + C_{rr} mgv + mv\frac{dv}{dt}\right) + P_{aux} \quad (19)$$

Regenerative braking should also be considered as it recovers part of the vehicle's kinetic energy as electrical energy during deceleration. The power through regenerative braking ($P_{regen}$) is derived from the kinetic energy of the vehicle:

$$P_{regen} = -\mu\left(mv\frac{dv}{dt}\right), \quad \frac{dv}{dt} < \beta \quad (20)$$

Where $\mu$ is regeneration braking efficiency, and $\beta \approx -0.045 \frac{m}{s^2}$ is regeneration braking deceleration threshold which can be found during the vehicle's deceleration state.

The complete dynamics, accounting for regenerative braking, is given by:

$$\mathcal{N}[P(v,t); \eta, \mu, m, C_{rr}, C_d,]$$
$$= \frac{1}{\eta}\left(\frac{1}{2}\rho A C_d v^3 + C_{rr} mgv + mv\frac{dv}{dt}(1 - \mu I_{a<\beta})\right) + P_{aux} \quad (21)$$

Where $I_{a<\beta}$ represents indicator function that activates $\mu$ based on regeneration breaking deceleration threshold $\beta$.

*D. Physics-Informed Neural Network Training*

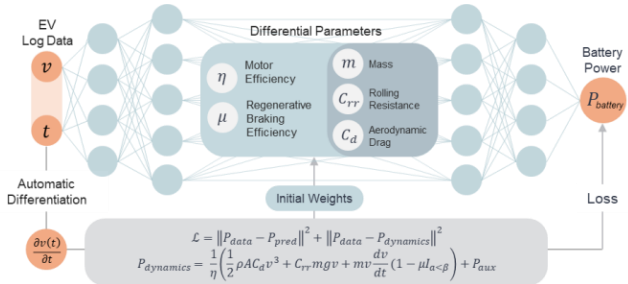

Fig. 4. PINN architecture

From the dynamics derived above, we can set-up the loss function which is crucial as the network not only should fit the vehicle log data but should also be constrained to the physics governing the vehicle's dynamics. Formulated similarly to the original PINN loss function, the overall loss function for EV Battery Power Consumption is expressed as:

$$\mathcal{L}_{power} = \sum_{i=1}^{N}\left(\|P_{data}^{(i)} - P_{pred}^{(i)}\|^2 + \lambda\|P_{physics}^{(i)} - P_{pred}^{(i)}\|^2\right) \quad (22)$$

$$P_{physics} = \mathcal{N}[P(v,t); \eta, \mu, m, C_{rr}, C_d] \quad (23)$$

Where $P_{data}, P_{pred}, P_{physics}, \lambda$ represents true battery power from log data, network's prediction, and theoretical battery power from dynamics, and weight, respectively.

By minimizing this combined loss function, the network not only learns to predict battery power based on vehicle speed but also estimates the key parameters governing the vehicle's dynamics. This approach helps prevent underfitting when data is sparse and mitigates overfitting by respecting the underlying physics. Incorporating the physics loss term adds local minima to the already complex, non-convex optimization landscape. To mitigate this, we initialized differential parameters with well-established values from online sources:

$$\begin{bmatrix} \eta \\ \mu \\ m \\ C_{rr} \\ C_d \end{bmatrix}_{initial,3LR} = \begin{bmatrix} 0.7 \\ 0.5 \\ 1823 \\ 0.0096 \\ 0.23 \end{bmatrix} \quad (24)$$

$$\begin{bmatrix} \eta \\ \mu \\ m \\ C_{rr} \\ C_d \end{bmatrix}_{initial,S} = \begin{bmatrix} 0.7 \\ 0.5 \\ 2250 \\ 0.0096 \\ 0.23 \end{bmatrix} \quad (25)$$

Regeneration efficiency ($\mu$) was given an initial guess of 0.5 as it was not available online. Vehicle mass and dimensions from the official Tesla Model 3 and S Owner's Manual [10], [11]. Other parameters such as air density were set at $\rho_{air} = 1.17 \frac{kg}{m^3}$ and frontal area was set at $A_{3LR} = 2.22 m^2$ and $A_S = 2.40 m^2$ respectively.

Instead of exploring the entire parameter space from scratch, we provided PINN with a realistic starting point, enabling faster convergence to accurate solutions.

The network was set up with vehicle speed and time as inputs and battery power as the output. After countless trials, a converging architecture was found to consist of four fully connected layers, each with 128 neurons, and with hyperbolic tangent (tanh) activation function. We also applied the Adam [12] optimizer with different learning rates assigned to the parameters and layers. We also determined the physics loss weighting factor to be $\lambda = 0.1$. Initially, the loss values were not converging properly, but by adjusting the learning rates, initializing parameters carefully, and fine-tuning the network architecture, both total loss and validation loss were reduced to approximately $1e - 3$.

TABLE II
PINN LOSS FOR TESLA MODEL 3 LONG RANGE

| Epoch | Data set* | Total Loss | Data Loss | Physics Loss |
|---|---|---|---|---|
| 1 | T | 35.527561 | 0.0277841 | 35.499778 |
|   | V | 11.830781 | 0.0308733 | 11.799907 |
| 10 | T | 4.3677359 | 0.0397093 | 4.3280265 |
|   | V | 2.8024990 | 0.0390067 | 2.7634923 |
| 100 | T | 0.1390878 | 0.0314255 | 0.1076622 |
|   | V | 0.1379689 | 0.0314022 | 0.1065666 |
| 1000 | T | 0.0219033 | 0.0115064 | 0.0103969 |
|   | V | 0.0182797 | 0.0115101 | 0.0067695 |
| 10000 | T | 0.0021949 | 0.0014924 | 0.0007025 |
|   | V | 0.0022917 | 0.0015515 | 0.0007402 |

*T represents Training data, V represents Validation data

TABLE III
PINN LOSS FOR TESLA MODEL S

| Epoch | Data set* | Total Loss | Data Loss | Physics Loss |
|---|---|---|---|---|
| 1 | T | 95.58700 | 0.0409975 | 95.54600 |
|   | V | 40.465814 | 0.0372440 | 40.428570 |
| 10 | T | 9.0055013 | 0.0258939 | 8.9796073 |
|   | V | 5.8529850 | 0.0266144 | 5.8263706 |
| 100 | T | 0.1297769 | 0.0270814 | 0.102695 |
|   | V | 0.1253943 | 0.0267145 | 0.0986798 |
| 1000 | T | 0.0364526 | 0.0175198 | 0.018932 |
|   | V | 0.0354756 | 0.0172076 | 0.0182680 |
| 10000 | T | 0.0035389 | 0.0015508 | 0.0019881 |
|   | V | 0.0033812 | 0.0013860 | 0.0019952 |

*T represents Training data, V represents Validation data

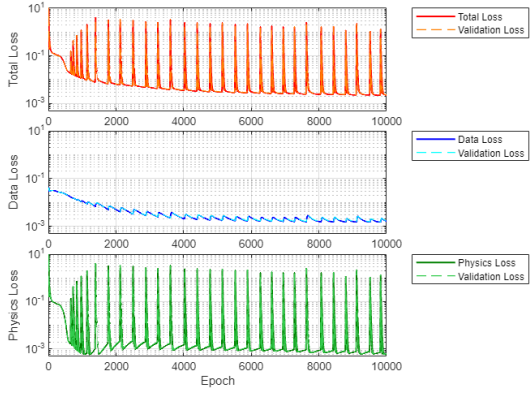

Fig. 5. PINN loss graph for Tesla Model 3 Long Range

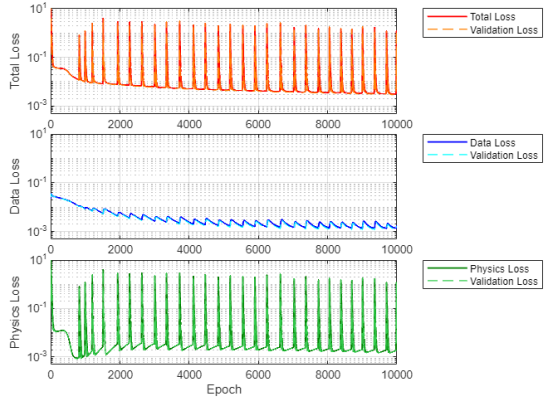

Fig. 6. PINN loss graph for Tesla Model S

*E. Runge-Kutta Neural Network Training*

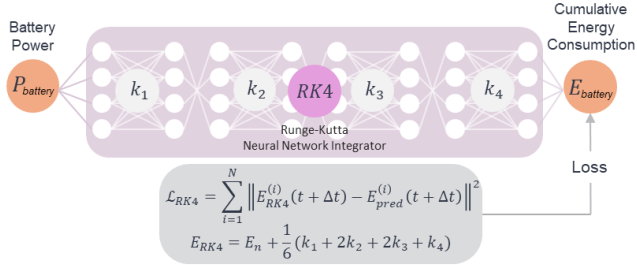

Fig. 7. RKNN architecture

To model the energy consumption of the EV, we need to integrate the power over time. Although directly utilizing energy data provided by the vehicle logs would have been ideal, our available data only included battery power. As exact integration of power dynamics is complex and computationally expensive to solve, we were inspired by "Runge-Kutta Neural Networks (RKNN) for Identification of Dynamical Systems in High Accuracy" [13] where Runge-Kutta subnetworks are embedded within a DNN to perform the Fourth Order Runge-Kutta. It demonstrated that predicting intermediate values is better than directly predicting the output value.

Implementation of RKNN can be simplified using PyTorch. In our implementation, each subnetwork sequentially learns the parameters $k_1, k_2, k_3, k_4$, similar to the classic Runge-Kutta method. We found that RKNNs offer more stable training compared to traditional DNNs, especially for deeper networks. RKNNs can achieve greater depth without these stability issues. We predict energy $E(t)$ at a future time $t + \Delta t$ using the following steps:

$$E_{RK4}(t + \Delta t) = E(t) + \frac{\Delta t}{6}(k_1 + 2k_2 + 2k_3 + k_4) \quad (25)$$

$$k_1 = \mathcal{N}_1(P(t)) \quad (26)$$

$$k_2 = \mathcal{N}_2\left(P\left(t + \frac{\Delta t}{2}\right), E(t) + \frac{\Delta t}{2}k_1\right) \quad (27)$$

$$k_3 = \mathcal{N}_3\left(P\left(t + \frac{\Delta t}{2}\right), E(t) + \frac{\Delta t}{2}k_2\right) \quad (28)$$

$$k_4 = \mathcal{N}_4(P(t + \Delta t), E(t) + \Delta t k_3) \quad (29)$$

Where $\mathcal{N}_i$ are Runge-Kutta subnetworks within the RKNN, $P$ represents Battery Power, and $\Delta t$ is time interval.

We can directly integrate RK4 method into a DNN to accept predicted battery power from PINN and use it to predict RK4 approximation of energy consumption with the following loss function.

$$\mathcal{L}_{RK4} = \sum_{i=1}^{N} \left\| E_{RK4}^{(i)}(t + \Delta t) - E_{pred}^{(i)}(t + \Delta t) \right\|^2 \quad (30)$$

We found that using 32 neurons with 3 hidden layers per subnetwork, and a learning rate of $1e-2$ achieved optimal stability and accuracy, reducing both total and validation losses to $2e-4$.

TABLE IV
RKNN PREDICTION LOSS

| Epoch | Data set* | RKNN Loss | DNN Loss |
|---|---|---|---|
| 1 | T | 0.9869467 | 1.0588666 |
|  | V | 0.9185590 | 0.9831669 |
| 10 | T | 0.3569548 | 0.2055948 |
|  | V | 0.3176739 | 0.2005833 |
| 100 | T | 0.0108608 | 0.0623520 |
|  | V | 0.0106572 | 0.0723817 |
| 1000 | T | 0.0024956 | 0.0063364 |
|  | V | 0.0026879 | 0.0059875 |
| 10000 | T | 0.0002584 | 0.0007926 |
|  | V | 0.0003183 | 0.0008287 |

*T represents Training data, V represents Validation data

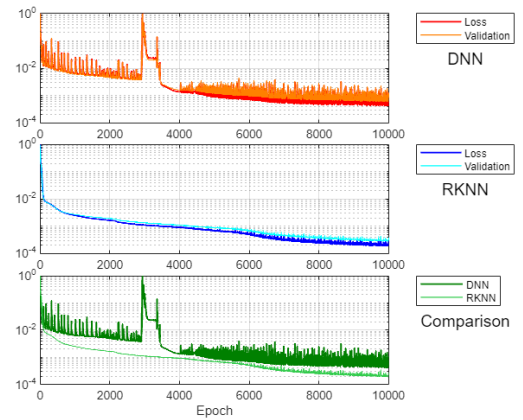

Fig. 8. RKNN loss graph

## IV. RESULTS

### A. Experimental Results

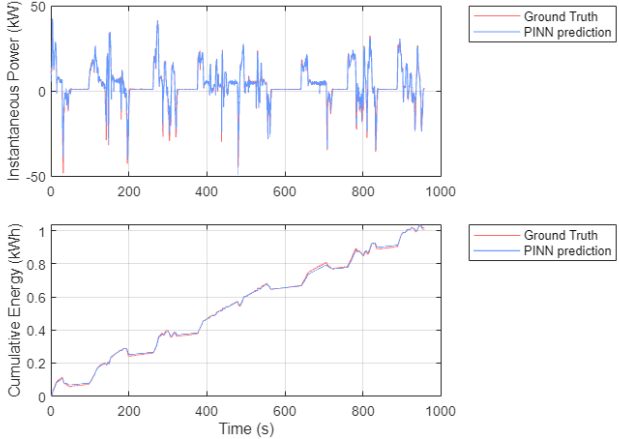

Fig. 9. Tesla Model 3 Battery Power and Energy Prediction

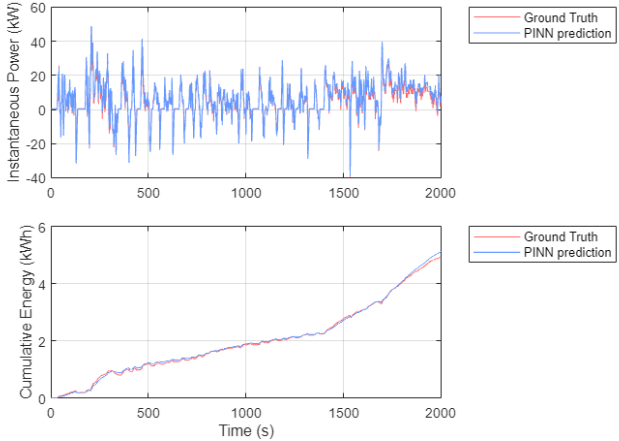

Fig. 10. Tesla Model S Battery Power and Energy Prediction

TABLE V
PREDICTED PARAMETERS

| Parameter | Vehicle | Initial Guess | Prediction | Absolute Error | Relative Error |
|---|---|---|---|---|---|
| $\eta$ | Model 3 LR | 0.7 | 0.7349 | 0.0349 | 4.98% |
|  | Model S |  | 0.7248 | 0.0248 | 3.54% |
| $\mu$ | Model 3 LR | 0.5 | 0.6743 | 0.1743 | 34.66% |
|  | Model S |  | 0.7038 | 0.2038 | 40.76% |
| $m$ | Model 3 LR | 1823 | 1975 | 152 | 8.33% |
|  | Model S | 2250 | 2313 | 63 | 2.8% |
| $C_{rr}$ | Model 3 LR | 0.0096 | 0.00915 | 0.00045 | 4.69% |
|  | Model S |  | 0.01100 | 0.00140 | 14.58% |
| $C_d$ | Model 3 LR | 0.23 | 0.2349 | 0.0049 | 2.13% |
|  | Model S |  | 0.2457 | 0.0157 | 6.83% |

TABLE VI
PARAMETER COMPARISON

| Vehicle | $\eta$ | $m$ | $C_{rr}$ | $C_d$ |
|---|---|---|---|---|
| BMW i4 | 0.75 | 2125 | 0.014 | 0.24 |
| Chevy Bolt | 0.72 | 1628 | 0.011 | 0.28 |
| Hyundai Ioniq 6 | 0.75 | 2086 | 0.0065 | 0.21 |
| Tesla Model 3 | 0.70 | 1823 | 0.0096 | 0.23 |
| PINN Prediction (Model 3 LR) | 0.7349 | 1975 | 0.00915 | 0.2349 |
| PINN Prediction (Model S) | 0.7248 | 2313 | 0.01100 | 0.2457 |

### B. Evaluation

Figure 9 and 10 demonstrate that the PINN predictions almost overlap the ground truth log data from the test vehicles. The predicted energy consumption also aligned with the RK4 integration of log data ground truth. Tables V and VI show the parameters predicted by the PINN, which align with the log data and converge to realistic values. Interestingly, the predicted mass for the Tesla Model 3 differs by 152 kg from the initial curb weight. It closely matches the combined weight of the two researchers and the cargo during the test drive.

## V. CONCLUSION

In this paper, we successfully applied Physics-Informed Neural Networks to predict EV power and energy consumption using only vehicle speed and time as inputs. The model not only achieved high accuracy in battery power prediction but also accurately estimated real-world parameters like vehicle mass and aerodynamic coefficients. These results highlight the potential of PINNs for enhancing EV energy management without relying on extensive sensor data. While the current study focused on specific vehicle models, future work could involve applying the model to a wider range of vehicles and driving conditions. Additionally, integrating the EV-PINN into real-time control could further optimize energy efficiency and path planning in EV

REFERENCES

[1] P. Patil, "The Future of Electric Vehicles: A Comprehensive Review of Technological Advancements, Market Trends, and Environmental Impacts," Journal of Artificial Intelligence and Machine Learning in Management, vol. 4, no. 1, pp. 56-68, 2020. [Online]. Available: https://journals.sagescience.org/index.php/jamm/article/view/63.

[2] E. Acar, S. Kardas, and M. Kumbasar, "A survey on design optimization of battery electric vehicle components, systems, and management," Journal of Cleaner Production, vol. 280, 2024, pp. 1-14. DOI: 10.1016/j.jclepro.2020.123456.

[3] W. Zhang and Z. Zhang, "Optimization and solution method for electric vehicle charging and discharging load," Energy Reports, vol. 7, 2021, pp. 45-58. DOI: 10.1016/j.egyr.2020.123456.

[4] C. Zhu, J. Wu, and X. Li, "Optimization strategies for real-time energy management of electric vehicles based on LSTM network learning," Energy Reports, vol. 8, 2022, pp. 1022-1033. DOI: 10.1016/j.egyr.2022.10.349.

[5] M. Raissi, P. Perdikaris, and G. E. Karniadakis, "Physics-informed neural networks: A deep learning framework for solving forward and inverse problems involving nonlinear partial differential equations," Journal of Computational Physics, vol. 378, 2019, pp. 686-707. DOI: 10.1016/j.jcp.2018.10.045.

[6] S. Park, C. Yun, J. Lee, and J. Shin, "A Survey on Universal Approximation Theorems," arXiv preprint arXiv:2407.12895, 2024. [Online]. Available: https://arxiv.org/abs/2407.12895.

[7] J. J. Craig, Introduction to Robotics: Mechanics and Control, 4th ed. Global Edition. Boston, MA: Pearson, 2022, ch. 6, pp. 200.

[8] J. Nicodemus, J. Kneifl, J. Fehr, and B. Unger, "Physics-informed Neural Networks-based Model Predictive Control for Multi-link Manipulators," IFAC-PapersOnLine, vol. 55, no. 20, pp. 331-336, 2022. doi: 10.1016/j.ifacol.2022.09.117.

[9] T. Kim, H. Lee and W. Lee, "Physics Embedded Neural Network Vehicle Model and Applications in Risk-Aware Autonomous Driving Using Latent Features," 2022 IEEE/RSJ International Conference on Intelligent Robots and Systems (IROS), Kyoto, Japan, 2022, pp. 4182-4189, doi: 10.1109/IROS47612.2022.9981303.

[10] Tesla Model 3 Owner's Manual, Tesla, Inc. [Online]. Available: https://www.tesla.com/ownersmanual/model3/en_us/

[11] Tesla Model S Owner's Manual, Tesla, Inc. [Online]. Available: https:// https://www.tesla.com/ownersmanual/models/en_us/

[12] D. P. Kingma and J. Ba, "Adam: A method for stochastic optimization," in *Proc. 3rd Int. Conf. Learn. Representations (ICLR)*, San Diego, CA, USA, May 2015, pp. 1-15.

[13] Yi-Jen Wang and Chin-Teng Lin, "Runge-Kutta neural network for identification of dynamical systems in high accuracy," in IEEE Transactions on Neural Networks, vol. 9, no. 2, pp. 294-307, March 1998, doi: 10.1109/72.661124.